%% file: main.tex
\documentclass{article}
\usepackage[square,sort,comma,numbers]{natbib}
\input{math_commands.tex}

\usepackage[margin=1in]{geometry}
\usepackage[T1]{fontenc}
\usepackage{url}
\usepackage{caption} 
\usepackage{graphicx}
\usepackage{algpseudocode,algorithm,algorithmicx}
\usepackage{booktabs}
\usepackage{tabularx}
\usepackage{authblk}

\newcommand{\feat}{f}
\newcommand{\sdist}[1]{(#1, c)\sim P_\sS^{#1,c}}
\newcommand{\udist}{c\sim P_\sU^{c}}
\newcommand{\ufdist}[1]{(#1, c)\sim P_\sU^{#1,c}}

\newcommand{\nei}{l^{sib}}
\newcommand{\fakefeat}{\Tilde{\feat}}
\newcommand{\predy}{\hat{y}}
\newcommand{\clf}{g}
\newcommand{\LsBCE}{\Ls_\texttt{BCE}}
\newcommand{\LsCLS}{\Ls_\texttt{CLS}}
\newcommand{\LsCYC}{\Ls_\texttt{CYC}}
\newcommand{\LsKEY}{\Ls_\texttt{KEY}}
\newcommand{\LsLDAM}{\Ls_\texttt{LDAM}}
\newcommand{\LsWGAN}{\Ls_\texttt{WGAN}}
\newcommand{\LsWGANZ}{\Ls_\texttt{WGAN-Z}}

\newcommand{\model}{AGMC-HTS}
\newcommand{\paraname}[1]{\noindent{\bf #1}}

\title{Generalized Zero-shot ICD Coding}

\author[$1$]{Congzheng Song\thanks{Email: cs2296@cornell.edu. Work done while the author was an intern at Petuum Inc}}
\author[$2$]{Shanghang Zhang}
\author[$2$]{Najmeh Sadoughi}
\author[$2$]{Pengtao Xie}
\author[$2$]{Eric Xing}

\affil[$1$]{Cornell University}
\affil[$2$]{Petuum Inc}

\date{}

\begin{document}

\maketitle            

\begin{abstract}
The International Classification of Diseases (ICD) is a list of classification codes for the diagnoses. Automatic ICD coding is in high demand as the manual coding can be labor-intensive and error-prone.
It is a multi-label text classification task with extremely long-tailed label distribution, making it difficult to perform fine-grained classification on both frequent and zero-shot codes at the same time. 
In this paper, we propose a latent feature generation framework for generalized zero-shot ICD coding, where we aim to improve the prediction on codes that have no labeled data without compromising the performance on seen codes. 
Our framework generates pseudo features conditioned on the ICD code descriptions and exploits the ICD code hierarchical structure. 
To guarantee the semantic consistency between the generated features and real features, we reconstruct the keywords in the input documents that are related to the conditioned ICD codes. To the best of our knowledge, this works represents the first one that proposes an adversarial generative model for the generalized zero-shot learning on multi-label text classification. 
Extensive experiments demonstrate the effectiveness of our approach. On the public MIMIC-III dataset, our methods improve the F1 score from nearly 0 to $20.91\%$ for the zero-shot codes, and increase the AUC score by 3\% (absolute improvement) from previous state of the art. We also show that the framework improves the performance on few-shot codes.

\end{abstract}

\input{intro.tex}
\input{related.tex}

\input{method.tex}

\input{experiment.tex}
\input{conclusion.tex}

\clearpage
\bibliographystyle{plain}
\bibliography{citation}

\clearpage
\appendix
\input{appendix.tex}

\end{document}

%% file: math_commands.tex
\usepackage{amsmath,amsfonts,bm}

\def\eqref#1{equation~\ref{#1}}
\def\Eqref#1{Equation~\ref{#1}}

\def\1{\bm{1}}

\DeclareMathAlphabet{\mathsfit}{\encodingdefault}{\sfdefault}{m}{sl}
\SetMathAlphabet{\mathsfit}{bold}{\encodingdefault}{\sfdefault}{bx}{n}

\def\gU{{\mathcal{U}}}
\def\gV{{\mathcal{V}}}

\def\sC{{\mathbb{C}}}

\def\sF{{\mathbb{F}}}

\def\sL{{\mathbb{L}}}

\def\sR{{\mathbb{R}}}
\def\sS{{\mathbb{S}}}

\def\sU{{\mathbb{U}}}

\def\sZ{{\mathbb{Z}}}

\newcommand{\E}{\mathbb{E}}
\newcommand{\Ls}{\mathcal{L}}

\newcommand{\rect}{\mathrm{rectifier}}
\newcommand{\softmax}{\mathrm{softmax}}
\newcommand{\sigmoid}{\sigma}



%% file: intro.tex
\section{Introduction}



In healthcare facilities, clinical records are classified into a set of International Classification of Diseases (ICD) codes that categorize diagnoses. 
ICD codes are used for a wide range of purposes including billing, reimbursement, and retrieving of diagnostic information.
Automatic ICD coding~\citep{stanfill2010systematic} is in great demand as manual coding can be labor-intensive and error-prone.
ICD coding is a multi-label text classification task, which is severely challenged by the following problems.
First, the distribution of the frequencies of ICD codes is highly long-tailed.
While some codes occur frequently, many other codes only have a few or even no labeled data due to the rareness of the disease. For example, in the medical dataset MIMIC III~\citep{johnson2016mimic}, among the 17,000 unique ICD-9 codes, 
more than $50\%$ of them never occur in the training data.
It is extremely challenging to perform fine-grained multi-label classification on both codes with labeled data (seen codes) and zero-shot (unseen) codes at the same time. 
Besides, clinical documents can be long and noisy, and extracting relevant information for all the codes need to be assigned  can be difficult. 
Automatic ICD coding for both seen and unseen codes fits into the generalized zero-shot learning (GZSL) paradigm~\citep{chao2016empirical}, where test data examples are from both seen and unseen classes and we classify them into the joint labeling space of both types of classes. Nevertheless, existing GZSL works focus on visual tasks~\citep{xian2017zero,liu2019large}. The study of GZSL for multi-label text classification is largely under-explored. In this work, we aim to bridge this gap.

%

\begin{figure}[t]
    \centering
    \includegraphics[width=0.8\textwidth]{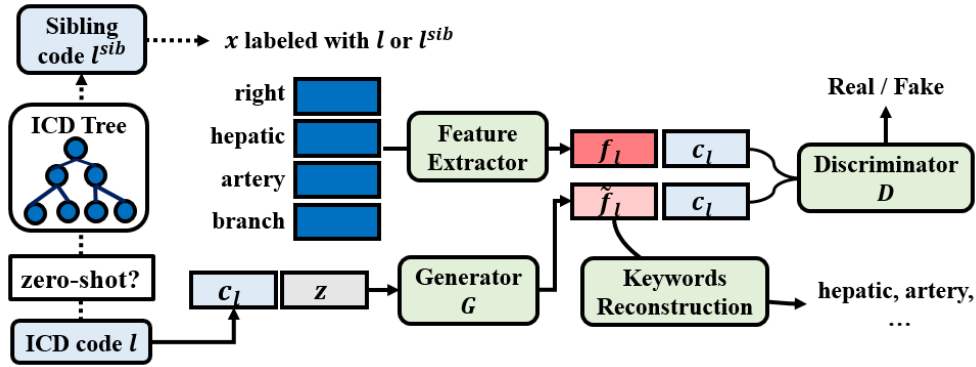}
    \caption{\footnotesize Overview of AGMC-HTS. The generator synthesizes features for an ICD code and the discriminator decides how realistic the input feature is. For a zero-shot ICD code, the discriminator distinguishes between the generated features and the real features from the data of its nearest sibling in the ICD hierarchy. The generated features are further used to reconstruct the keywords in the input documents to preserve semantics.} 
    \label{fig:overview}
\end{figure}

To tackle the problem of generalized zero-shot ICD coding, we propose AGMC-HTS, an \textbf{A}dversarial \textbf{G}enerative \textbf{M}odel \textbf{C}onditioned on ICD code descriptions to generate pseudo examples in the latent feature space by exploiting the ICD code \textbf{H}ierarchical \textbf{T}ree \textbf{S}tructure.
Specifically, as illustrated in Figure \ref{fig:overview}, AGMC-HTS consists of a generator to synthesize code-specific latent features based on the ICD code descriptions, and a discriminator to decide how realistic the generated features are.
To guarantee the semantic consistency between the generated features and real features, AGMC-HTS reconstructs the keywords in the input documents that are related to the conditioned ICD codes.
Such a pseudo cycle generation architecture especially benefits the feature generation of zero-shot codes. 
Different from the pure cycle architecture, we only generate the keywords instead of the whole text, which significantly eases the training of the model and adds more semantics to the synthetic features. 
To further facilitate the feature synthesis of zero-shot codes, we take advantage of the hierarchical structure of the ICD codes and encourage the zero-shot codes to generate similar features with their nearest sibling code. 
Besides ICD coding, the proposed AGMC-HTS can be applied to various text classification problems,
such as indexing biomedical articles and patent classification.

The contributions of this paper are summarized as follows:
1) To the best of our knowledge, this work represents the first one that proposes an adversarial generative model for the generalized zero-shot learning on multi-label text classification.
AGMC-HTS generates pseudo document features conditioned on the zero-shot codes and finetunes the ICD code assignment classifier. 
2) AGMC-HTS incorporates the hierarchical structure and domain knowledge of the codes to ensure the semantic relevance between the latent features and codes. 
3) We propose a pseudo cycle generation architecture to guarantee the semantic consistency between the generated features and real features by reconstructing the keywords extracted from real input texts. 
It also benefits the feature generation of zero-shot codes, which do not have document samples in the training data.
4) Extensive experiments demonstrate the effectiveness of our approach. On the public MIMIC-III dataset, our methods improve the F1 score from nearly 0 to $20.91\%$ for the zero-shot codes, and increase the AUC score by 3\% (absolute improvement) from previous state of the art. We also show that the framework improves the performance on few-shot codes with a handful of labeled data.

%% file: related.tex
\section{Related Work}
\paraname{Automated ICD coding.} Several approaches have explored automatic assigning ICD codes on clinical text data~\citep{stanfill2010systematic}. 
\cite{mullenbach2018explainable} proposed to extract per-code textual features with attention mechanism for ICD code assignments. \cite{shi2017towards} explored character based short-term memory (LSTM) with attention and \cite{xie2018neural} applied tree LSTM with ICD hierarchy information for ICD coding. 
Most existing work either focused on predicting the most common ICD code or did not utilize the ICD hierarchy structure for prediction.  
\cite{rios2018few} proposed a neural network models incorporating ICD hierarchy information that improved the performance on the rare and zero-shot codes. 
The performance is evaluated in terms of the relative ranks to other infrequent codes
The model hardly ever assign rare codes in its final prediction as we show in Section~\ref{sec:result}, making it impractical to deploy in real applications.

\paraname{Feature generation for GZSL.} 
The idea of using generative models for GZSL is to generate latent features for unseen classes and train a classifier on the generated features and real features for both seen and unseen classes. \cite{xian2018feature} proposed using conditional GANs to generate visual features given the semantic feature for zero-shot classes. 
\cite{felix2018multi} added a cycle-consistent loss on generator to ensure the generated features captures the class semantics by using linear regression to map visual features back to class semantic features. \cite{ni2019dual} further improves the semantics preserving using dual GANs formulation instead of a linear model. Previous works focus on vision domain where the features are extracted from well-trained deep models on large-scale image dataset. 
We introduce the first feature generation framework tailored for zero-shot ICD coding by exploiting existing medical knowledge from limited available data.

\paraname{Zero-shot text classification.}
\cite{pushp2017train} has explored zero-shot text classification by learning relationship between text and weakly labeled tags on large corpus. The idea is similar to~\cite{rios2018few} in learning the relationship between input and code descriptions. 
\cite{zhangkumjornZeroShot} introduced a two-phase framework for zero-shot text classification. An input is first 
determined as from a seen or an unseen classes before the final classification. This approach does not directly apply to ICD coding as the input is labeled with a set of codes which can include both seen and unseen codes 
It is not possible to determine if the data is from a seen or an unseen class.

%% file: method.tex
\section{Method}

The task of automatic ICD coding is to assign ICD codes to patient's clinical notes. We formulate the problem as a multi-label text classification problem.
Let $\sL$ be the set of all ICD codes and $L = |\sL|$, given an input text, the goal is to predict $y_l\in\{0, 1\}$ for all $l\in\sL$. Each ICD code $l$ has a short text description. For example, the description for ICD-9 code 403.11 is
\emph{``Hypertensive chronic kidney disease, benign, with chronic kidney disease stage V or end stage renal disease."} There is also a known hierarchical tree structure on all the ICD codes: for a node
representing an ICD code, the children of this node
represent the subtypes of this ICD code.

We focus on the generalized zero-shot ICD coding problem: accurately assigning code $l$ given that $l$ is never assigned to any training text (i.e. $y_l=0$), without sacrificing the performance on codes with training data.
We assume a pretrained model as a feature extractor that performs 
ICD coding by extracting label-wise feature $\feat_l$ and predicting $y_l$ by $\sigmoid(\clf_l^\top\cdot\feat_l)$, where $\sigmoid$ is the sigmoid function and $\clf_l$ is the binary classifier for code $l$. 
For the zero-shot codes, $\clf_l$ is never trained on $\feat_l$ with $y_l=1$ and thus at inference time, the pretrained feature extractor hardly ever assigns zero-shot codes. 

Figure~\ref{fig:overview} shows an overview of our method. We propose to use generative adversarial networks (GAN)~\citep{goodfellow2014generative} to generate $\fakefeat_l$ with $y_l=1$ by conditioning on code $l$. The generator $G$ tries to generate the fake feature $\fakefeat$ given an ICD code description. The discriminator $D$ tries to distinguish between $\fakefeat$ and real latent feature $\feat$ from the feature extractor model.  After the GAN is trained, we use $G$ to synthesize $\fakefeat_l$ and fine-tune the binary classifier $\clf_l$ with $\fakefeat_l$ for a given zero-shot code $l$. Since the binary code classifiers are independently fine-tuned for zero-shot codes, the performance on the seen codes is not affected, achieving the goal of GZSL.

\subsection{Feature extractor}
\label{sec:base}
\begin{figure}[t]
    \centering
    \includegraphics[width=0.8\textwidth]{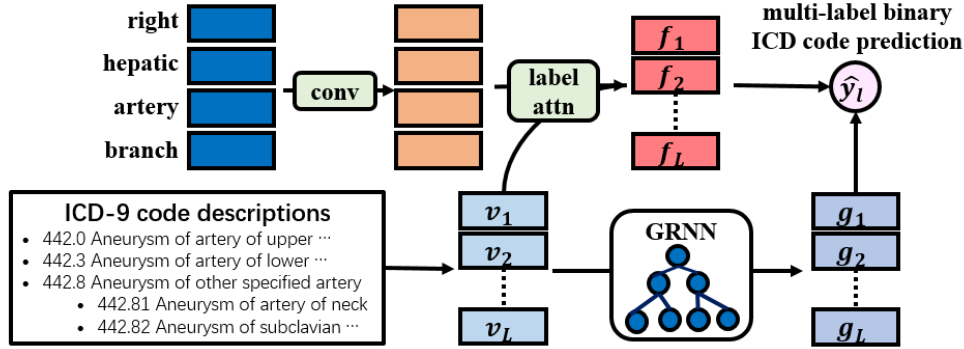}
    \caption{\footnotesize ZAGRNN as the feature extractor model . ZAGRNN extracts label-wise features and construct embedding of each ICD codes using GRNN. ZAGRNN makes a binary prediction for each code based on the  dot product between graph label embedding and the label specific feature.}
    \label{fig:base}
\end{figure}

We first describe the feature extractor model that will be used for training the GAN.
The model is zero-shot attentive graph recurrent neural network (ZAGRNN), modified from the only previous work (to the best of our knowledge) that is tailored towards solving zero-shot ICD coding~\citep{rios2018few}. Figure~\ref{fig:base} shows the architecture of the ZAGRNN. At a high-level, given an input $x$, ZAGRNN extracts label-wise feature $\feat_l$ and performs binary prediction on $\feat_l$ for each ICD code $l$.

\paraname{Label-wise feature extraction.}
Given an input clinical document $x$ containing $n$ words, we represent it with a matrix $X = [w_1, w_2, \dots, w_n]$ where $w_i\in\sR^{d}$ is the word embedding vector for the $i$-th word.  Each ICD code $l$ has a textual description.
To represent $l$, we construct an embedding vector $v_l$ by averaging the embeddings of words in the description. 

The word embedding is shared between input and label descriptions for sharing learned knowledge.
Adjacent word embeddings are combined using a one-dimension convolutional
neural network (CNN) to get the n-gram text features $H = \mathrm{conv}(X) \in\sR^{N\times d_c}$.
 Then the label-wise attention feature $a_l\in \sR^{d}$ for label $l$ is computed by:
\begin{align*}
    s_l &= \softmax(\mathrm{tanh}(H\cdot W_a^\top + b_a)\cdot v_l) \quad \text{ for } l = 1, 2, \dots L\\
    a_l &= s_l^\top\cdot H  \quad \text{ for } l = 1, 2, \dots L
\end{align*}
where $s_l$ is the attention scores for all rows in $H$ and $a_l$ is the attended output of $H$ for label $l$. Intuitively, $a_l$ extracts the most relevant information in $H$ about the code $l$ by using attention. Each input then has in total $L$ attention feature vectors for each ICD code.

\paraname{Multi-label classification.} For each code $l$, the binary prediction $\predy_l$ is generated by:
\begin{align*}
\feat_l &= \rect( W_o\cdot a_l + b_o),
\qquad
\hat{y}_l=\sigmoid(\clf_l^\top\cdot \feat_l)
\end{align*}
We use graph gated recurrent neural networks (GRNN)~\citep{li2015gated} to encode the classifier $\clf_l$. Let $\gV(l)$ denote the set of adjacent codes of $l$ from the ICD tree hierarchy and $t$ be the number of times we propagate the graph, the classifier $\clf_l=\clf_l^t$ is computed by: 
\begin{align}
\label{eq:grnn}
    \clf^0_l=v_l, \quad h^{t}_l = \frac{1}{|\gV(l)|}\Sigma_{j\in\gV(l)}\clf_j^{t-1},\quad \clf_l^t = \mathrm{GRUCell}(h^t_l, \clf_l^{t-1})
\end{align}
where $\mathrm{GRUCell}$ is a gated recurrent units~\citep{chung2014empirical} and the construction is detailed in Appendix~\ref{apx:gru}. The weights of the binary code classifier is tied with the graph encoded label embedding $g_l$ so that the learned knowledge can also benefit zero-shot codes since label embedding is computed from a shared word.

The loss function for training is multi-label binary cross-entropy:
\begin{align}
    \mathcal{L}_\texttt{BCE}(y, \predy) = -\sum_{l=1}^L[y_l\log(\hat{y_l}) + (1-y_l)\log(1 - \hat{y_l})]
\end{align}

As mentioned above, the distribution of ICD codes is extremely long-tailed. To counter the label imbalance issue, we adopt label-distribution-aware margin (LDAM)~\citep{cao2019learning}, where we subtract the logit value before sigmoid function by a label-dependent margin $\Delta_l$:
\begin{align}
    \predy^m_l &= \sigmoid(\clf_l^\top\cdot \feat_l - \mathbf{1}(y_l=1)\Delta_l)
\end{align}
where function $\mathbf{1}(\cdot)$ outputs 1 if $y_1=1$ and $\Delta_l = \frac{C}{n_l^{1/4}}$ and $n_l$ is the number of training data labeled with $l$ and $C$ is a constant. The LDAM loss is thus: $\LsLDAM = \LsBCE(y, \predy^m)$.

\subsection{Zero-shot Latent Feature Generation with WGAN-GP}
For a zero-shot code $l$, the code label $y_l$ for any training data example is $y_l=0$ and the binary classifier $\clf_l$ for code assignment is never trained with data examples with $y_l=1$ due to the dearth of such data. Previous works have successfully applied GANs for GZSL in the vision domain~\citep{xian2018feature,felix2018multi}. We propose to use GANs to improve zero-shot ICD coding by generating pseudo data examples in the latent feature space for zero-shot codes and fine-tuning the code-assignment binary classifiers using the generated latent features.  

More specifically, we use the Wasserstein GAN~\citep{arjovsky2017wasserstein} with gradient penalty (WGAN-GP)~\citep{gulrajani2017improved}  to generate 
code-specific latent features conditioned on the textual description of each code. Detail of WGAN-GP is described in Appendix~\ref{apx:related}. To condition on the code description, we use a label encoder function $C: \sL\mapsto\sC$ that maps the code description to a low-dimension vector $c$. We denote $c_l = C(l)$. The generator, $G:\sZ\times\sC \mapsto \sF$, takes in a random Gaussian noise vector $z\in \sZ$ and an encoding vector $c\in\sC$ of a code 
description to generate a latent feature $\fakefeat_l=G(z,c)$ for this code. The discriminator or 
critic, $D:\sF\times\sC\mapsto\sR$, takes in a latent feature vector $\feat$ (either generated by WGAN-GP or 
extracted from real data examples) and the encoded label vector $c$ to produce a real-valued score $D(\feat, c)$ representing how realistic $\feat$ is. The WGAN-GP loss is:
\begin{align}
\label{eq:wgan}
    \mathcal{L}_\texttt{WGAN} =& \E_{\sdist{\feat}}[D(\feat, c))] - \E_{\sdist{\fakefeat}}[D(\fakefeat, c))] + \nonumber 
    \\ & \lambda\cdot \E_{\sdist{\hat{\feat}}}[(||\nabla D(\hat{\feat}, c))||_2 -1)^2]
\end{align}
where $\sdist{\cdot}$ is the joint distribution of latent features and encoded label vectors from the set of seen code labels $\sS$,
$\hat{\feat}=\alpha\cdot\feat + (1-\alpha)\cdot\fakefeat$ with $\alpha\sim\gU(0, 1)$ and $\lambda$ is the gradient penalty coefficient. WGAN-GP can be learned by solving the minimax problem: $\min_G \max_D \LsWGAN$.

\paraname{Label encoder.} The function $C$ is an ICD-code encoder that maps a code description to an embedding vector. For a code $l$, we first use a LSTM~\citep{hochreiter1997long} to encode the sequence of $M$ words in the description into a sequence of hidden states $[e_1, e_2, \dots, e_M]$. We then perform a dimension-wise max-pooling over the hidden state sequence to get a fixed-sized encoding vector $e_l$. Finally, we obtain the eventual embedding $c_l=e_l||g_l$ of code $l$ by concatenating $e_l$ with $\clf_l$ which is the embedding of $l$ produced by the graph encoding network. $c_l$ contains both the latent semantics of the description (in $e_l$) as well as the ICD hierarchy information (in $\clf_l$).

\paraname{Keywords reconstruction loss.} To ensure the generated feature vector $\fakefeat_l$ captures the semantic meaning of code $l$, we encourage $\fakefeat_l$ to be able to well reconstruct the keywords extracted from the clinical notes associated with code $l$.  

For each input text $x$ labeled with code $l$, we extract the label-specific keyword set $K_l=\{w_1, w_2, \dots,w_k\}$ as the set of most similar words in $x$ to $l$, where the similarity is measured by cosine similarity between word embedding in $x$ and label embedding $v_l$. Let $Q$ be a projection matrix, $\mathcal{K}$ be the set of all keywords from all inputs and $\pi(\cdot,\cdot)$ denote the cosine similarity function, the loss for reconstructing keywords given the generated feature is as following:
\begin{align}
\label{eq:key}
    \LsKEY &= -\log P(K_l |\Tilde{\feat}_l )\approx -\sum_{w_k\in K_l}\pi(w_k, v_l)\cdot\log P(w_k|\fakefeat_l) \nonumber \\
    &= -\sum_{w_k\in K_l}\pi(w_k, v_l)\cdot\log \frac{\exp(w_k^\top\cdot Q\fakefeat_l)}{\sum_{w\in\mathcal{K}} \exp (w^\top\cdot Q\fakefeat_l)}
\end{align}

\paraname{Discriminating zero-shot codes using ICD hierarchy.} In the current WGAN-GP framework, the discriminator cannot be trained on zero-shot codes due to the lack of real positive features. In order to include zero-shot codes during training, we utilize the ICD hierarchy and use $\feat^{sib}$, the latent feature extracted from real data of the nearest sibling $\nei$ of a zero-shot code $l$, for training the discriminator. This formulation would encourage the generated feature $\fakefeat$ to be close to the real latent features of the siblings of $l$ and thus $\fakefeat$ can better preserving the ICD hierarchy.  More formally, let $c^{sib}=C(\nei)$, we propose the following modification to $\LsWGAN$ for training zero-shot codes:
\begin{align}
    \LsWGANZ =&  
    \E_{\udist}[\pi(c, c^{sib}) \cdot  D(\feat^{sib}, c)] - \E_{\ufdist{\fakefeat}}[\pi(c, c^{sib}) \cdot D(\fakefeat, c)] + \nonumber 
    \\& \lambda\cdot \E_{\ufdist{\hat{\feat}}}[(||\nabla D(\hat{\feat}, c)||_2 -1)^2]
\end{align}
where $\udist$ is the distribution of encoded label vectors for the set of zero-shot codes $\sU$ and $\ufdist{\cdot}$ is defined similarly as in \Eqref{eq:wgan}. 
The loss term by the cosine similarity $\pi(c,c^{sib})$ to 
prevent generating exact nearest sibling feature for the 
zero-shot code $l$. After adding zero-shot codes to training, our full learning objective becomes:
\begin{align}
    \min_G\max_D \LsWGAN + \LsWGANZ + \beta\cdot\LsKEY
\end{align}
where $\beta$ is the balancing coefficient for keyword reconstruction loss.

\paraname{Fine-tuning on generated features.} After WGAN-GP is trained, we fine-tune the pretrained classifier $\clf_l$ from baseline model with generated features for a given zero-shot code $l$. We use the generator to synthesize a set of $\fakefeat_l$  and label them with $y_l=1$ and collect the set of $\feat_l$ from training data with $y_l=0$ using baseline model as feature extractor. We finally fine-tune $\clf_l$ on this set of labeled feature vectors to get the final binary classifier for a given zero-shot code $l$.

%% file: experiment.tex
\section{Experiments}
\subsection{Setup}

\paraname{Dataset description.}
We use the publicly available medical dataset MIMIC-III~\citep{johnson2016mimic} for evaluation, which contains approximately 58,000 hospital admissions of 47,000 patients who stayed in the ICU of the Beth Israel Deaconess Medical Center between 2001 and 2012. Each admission record has a discharge summary that includes medical history, diagnosis outcomes, surgical procedures, discharge instructions, etc. Each admission record is assigned with a set of most relevant ICD-9 codes by medical coders. The dataset is preprocessed as in~\citep{mullenbach2018explainable}. Our goal is to accurately predict the ICD codes given the discharge summary.

We split the dataset for training, validation, and testing by patient ID. In total we have 46,157 discharge summaries for training, 3,280 for validation and 3,285 for testing. There are 6916 unique ICD-9 diagnosis codes in MIMIC-III and 6090 of them exist in the training set. We use all the codes for training while using codes that have more than 5 data examples for evaluation. 
There are 96 out of 1,646 and 85 out of 1,630 unique codes  are zero-shot codes in validation and test set,
respectively.

\paraname{Baseline methods.} We compare our method with previous state of the art approaches on zero-shot ICD coding~\citep{rios2018few} as described in Section~\ref{sec:base}, meta-embedding for long-tailed problem~\citep{liu2019large} and WGAN-GP with classification loss $\LsCLS$~\citep{xian2018feature} and with cycle-consistent loss $\LsCYC$~\citep{felix2018multi} that were applied to ZSL in computer vision domain.  Detailed description and hyper-parameters of baseline methods are in Appendix~\ref{appendix:base}.

\paraname{Training details.} For WGAN-GP based methods, the real latent features are extracted from the final layer in the ZAGRNN model. Only features $\feat_l$ for which $y_l=1$ are collected for training.
We use a single-layer fully-connected network with hidden size 800 for both generator and discriminator.
For the code-description encoder LSTM, we set the hidden size to 200. We train the discriminator 5 iterations per each generator training iteration.
We optimize the WGAN-GP with ADAM~\citep{kingma2014adam} with mini-batch size 128 and learning rate 0.0001.
We train all variants of WGAN-GP for 60 epochs. We set the weight of $\LsCLS$ to 0.01 and $\LsCYC, \LsKEY$ to 0.1. For $\LsKEY$, we predict the top 30 most relevant keywords given the generated features.

After the generators are trained, we synthesize 256 features for each zero-shot code $l$ and fine-tune the classifier $g_l$ using ADAM and set the learning rate to 0.00001 and the batch size to 128.  We fine-tune on all zero-shot codes and select the best performing model on validation set and evaluate the final result on the test set.

\begin{table}[t]
    \centering
    \footnotesize
    \caption{\footnotesize Baseline ICD coding results on all the codes.}
    \begin{tabular}{l|rrrr|rrrr}
    \toprule 
     & \multicolumn{4}{c|}{Micro} & \multicolumn{4}{c}{Macro} \\
    Method &  Pre & Rec & F1 & AUC & Pre & Rec & F1 & AUC \\
    \midrule
    ZAGRNN~\citep{rios2018few} & 58.06 & 44.94 & 50.66 & 96.67 & 30.91 & 25.57 & 27.99 & 94.03 \\
    ZAGRNN + $\mathcal{L}_\texttt{LDAM}$~\citep{cao2019learning} & 56.06 & 47.14 & 51.22 & 96.70 & 31.72 & 28.06 & 29.78 & 94.08 \\
    \bottomrule
    \end{tabular}
    \label{tab:allshot}
\end{table}

\input{figures/zeroshot_tab.tex}

\subsection{Results}
\label{sec:result}
 We report both the micro and macro precision, recall, F1 and AUC scores on the zero-shot codes for all methods. Micro metrics aggregate the contributions of all codes to compute the average score while macro metrics compute the metric independently for each code and then take the average.
 All scores are averaged over 10 runs using different random seeds.
 
 Table~\ref{tab:allshot} shows the results of ZAGRNN models on all the code. Note that fine-tuning the zero-shot codes classifier using meta-embedding or WGAN-GP will not affect the classification for seen codes since the code assignment classifiers are independently fine-tuned.
 
 Table~\ref{tab:zeroshot} summarizes the results for zero-shot codes. 
 For the baseline ZAGRNN and meta-embedding models, the AUC on zero-shot codes is much better than random guessing. $\LsLDAM$ improves the AUC scores and meta-embedding can achieve slighter better F1 scores. 
 However, since these methods never train the binary classifiers for zero-shot codes on positive examples, both micro and macro recall and F1 scores are close to zero.
 In other words, these models almost never assign zero-shot codes at inference time. 
 For WGAN-GP based methods, all the metrics improve from ZAGRNN and meta-embedding except for micro precision. 
 This is due to the fact that the binary zero-shot classifiers are fine-tuned on positive generated features which drastically increases the chance of the models assigning zero-shot codes.

\paraname{Ablation studies on WGAN-GP methods.} 
We next examine the detailed performance of WGAN-GP methods using different losses.
Adding $\LsCLS$ hurts the micro metrics, which might be counter-intuitive at first. 
However, since the  $\LsCLS$ is computed based on the pretrained classifiers, which are not well-generalized on infrequent codes, adding the loss might provide bad gradient signal for the generator.
Adding $\LsCYC, \LsKEY$ and $\LsWGANZ$ improves $\LsWGAN$ and achieves comparable performances in terms of both micro and macro metrics. 
At a closer look, $\LsWGANZ$ improves the recall most, which matches the intuition that learning with the sibling codes enables the model to generate more diverse latent features.
The performance drops when combing  $\LsWGANZ$ with $\LsCLS$ and $\LsCYC$. 
We suspect this might be due to a conflict of optimization that the generator tries to synthesize $\fakefeat$ close to the sibling code $\nei$ and simultaneously maps $\fakefeat$ back to the exact code semantic space of $l$. 
Using $\LsKEY$ resolves the conflict as it reconstructs more generic semantics from the
words instead of from the exact code descriptions.
Our final model that uses the combination of $\LsWGANZ$ and $\LsKEY$ achieves the best performance on both micro and macro F1 and AUC score.

\begin{figure}[t]
    \centering
    \begin{tabular}{cc}
 \includegraphics[width=0.4\textwidth]{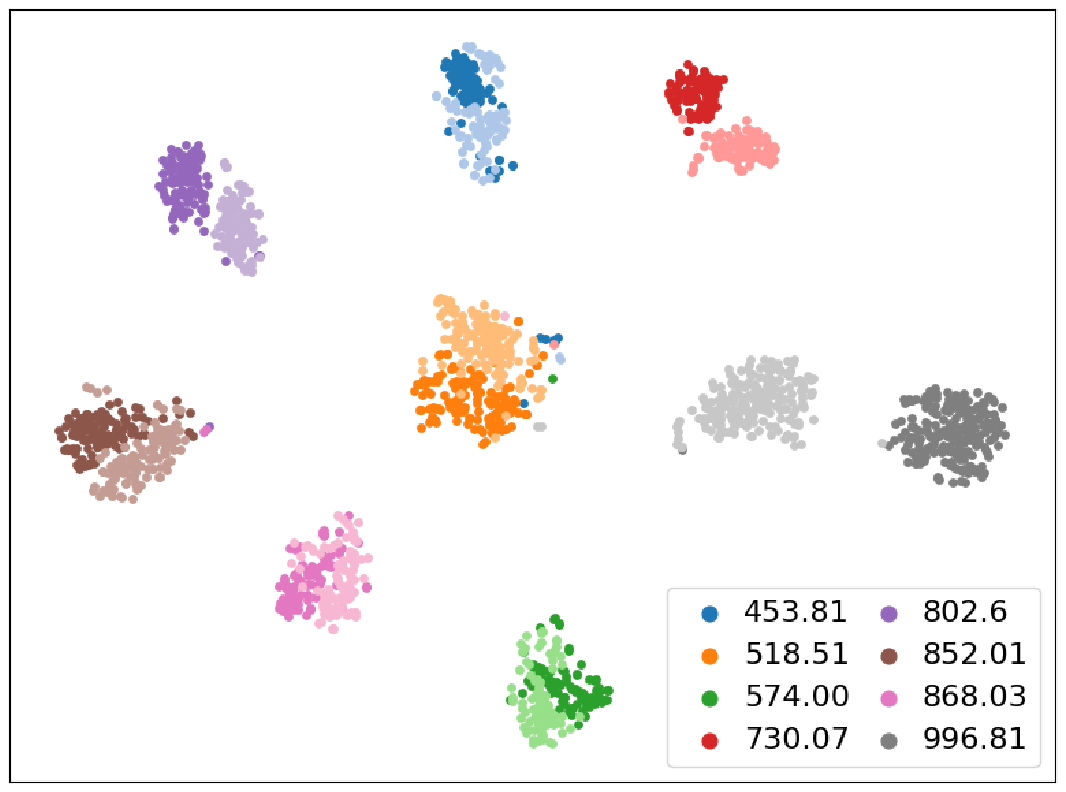}
    &     
\includegraphics[width=0.4\textwidth]{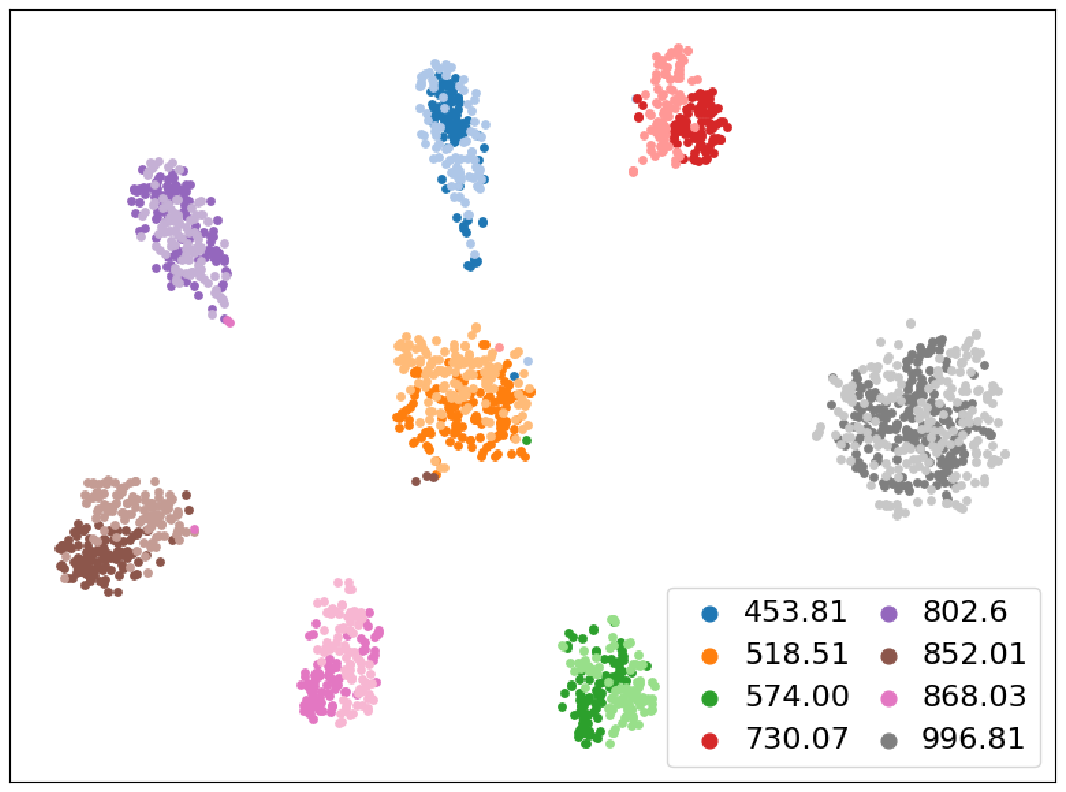}
\\
(a) $\LsWGAN$         &  (b) $\LsWGANZ$
    \end{tabular}
    \caption{\footnotesize T-SNE projection of generated features for zero-shot codes using (a) $\LsWGAN$ and (b) $\LsWGANZ$. Lighter color are projection of generated features and darker color are of real features from the nearest sibling codes.}
    \label{fig:tsne}
\end{figure}

\paraname{T-SNE visualization of generated features.}
We plot the T-SNE projection of the generated features for zero-shot codes using WGAN-GP with $\LsWGAN$ and $\LsWGANZ$ in Figure~\ref{fig:tsne}.
Dots with lighter color represent the  projections of generated features and those with darker color correspond to the real features from the nearest sibling codes. Features generated for zero-shot codes using $\LsWGANZ$ are closer to the real features from the nearest sibling codes.  This shows that using $\LsWGANZ$ can generate features that better preserve the ICD hierarchy. 

\input{figures/keywords_tab.tex}
\paraname{Keywords reconstruction from generated features.}
We next qualitatively evaluate the generated features by examining their reconstructed keywords. 
We first train a keyword predictor using $\LsKEY$ on the real latent features and their keywords extracted from training data. Then we feed the generated features from zero-shot codes into the keyword predictor to get the reconstructed keywords.

Table~\ref{tab:keywords} shows some examples of the top predicted keywords for zero-shot codes. Even the keyword predictor is never trained on zero-shot code features, the generated features are able to find relevant words that are semantically close to the code descriptions. 
In addition, features generated with $\LsWGANZ$ can find more relevant keywords than $\LsWGAN$. 
For instance, for zero-shot code V10.62, the top predicted keywords from $\LsWGANZ$ include \emph{leukemia, myelogenous, CML (Chronic myelogenous leukemia)} which are all related to myeloid leukemia, a type of cancer of the blood and bone marrow.

\input{figures/fewshot_tab.tex}
\paraname{Few-shot codes results.} 
As we have seen promising results on zero-shot codes, we also evaluate our feature generation framework on few-shot ICD codes, where the number of training data for such codes are less than or equal to 5. We apply the exact same setup as zero-shot codes for synthesizing features and fine-tuning classifiers for few-shot codes. There are 220 and 223 few-shot codes in validation and test set,
respectively.

Table~\ref{tab:fewshot} summarizes the results. 
The performance of ZAGRNN models on few-shot codes is slightly better than zero-shot codes yet the recall are still very low. 
Meta-embedding can boosts the recall and F1 scores from baseline models. 
WGAN-GP methods can further boosts the performance on recall, F1 and AUC scores and the performance using different combination of losses generally follows the pattern in zero-shot code results.
In particular, $\LsWGANZ$ and $\LsKEY$ can perform slightly better than other WGAN-GP models in terms of F1 and AUC scores. 

%% file: figures/zeroshot_tab.tex
\begin{table}[t]
    \centering
    \footnotesize
    \caption{\footnotesize Zero-shot ICD coding results. Scores are averaged over 10 runs on different seeds.}
    \begin{tabular}{l|rrrr|rrrr}
    \toprule 
     & \multicolumn{4}{c|}{Micro} & \multicolumn{4}{c}{Macro} \\
    Method &  Pre & Rec & F1 & AUC & Pre & Rec & F1 & AUC \\
    \midrule
    ZAGRNN~\citep{rios2018few} & 0.00 & 0.00 & 0.00 & 89.05 & 0.00 & 0.00 & 0.00 & 90.89\\
    ZAGRNN + $\LsLDAM$~\citep{cao2019learning} & 0.00 & 0.00 & 0.00 & 90.78 & 0.00 & 0.00 & 0.00 & 91.91 \\
    ZAGRNN + Meta~\citep{liu2019large} & \textbf{46.70} & 0.89 & 1.74 & 90.08 & 3.88 & 0.95 & 1.52 & 91.88
    \\
    \midrule
    $\LsWGAN$~\citep{xian2018feature} & 23.92 & 17.63 & 20.30 & 91.94 & 17.30 & 17.38 & 17.34 & 92.26 \\
    $\LsWGAN$ +  $\LsCLS$ ~\citep{xian2018feature} & 23.57 & 16.55 & 19.44 & 91.71 & \textbf{18.39} & 16.81 & 17.56 & 92.32 \\ 
    $\LsWGAN$ +  $\LsCYC$~\citep{felix2018multi}& 23.97 & 17.93 & 20.51 & 91.88 & 17.86 & 17.83 & 17.84 & 92.27
    \\ 
    \midrule
    $\LsWGANZ$ + $\LsCLS$ & 22.49 & 17.40 & 19.62 & 91.80 & 16.56 & 17.26 & 16.90 & 92.16 \\
    $\LsWGANZ$ +  $\LsCYC$ & 21.44 & 17.24 & 19.11 & 91.90 & 16.05 & 17.06 & 16.54 & 92.25
    \\ \midrule
    $\LsWGAN$ +  $\LsKEY$ (Ours) & 23.26 & 18.24 & 20.45 & 91.73 & 17.09 & 18.38 & 17.71 & 92.21 \\ 
    $\LsWGANZ$ (Ours) & 22.18 & 19.03 & 20.48 & 91.79 & 16.87 & 18.84 & 17.80 & 92.28 \\ 
    $\LsWGANZ$ +  $\LsKEY$ (Ours) & 22.54 & \textbf{19.51} & \textbf{20.91} & \textbf{92.18} & 17.70 & \textbf{19.15} & \textbf{18.39} & \textbf{92.34} \\
    \bottomrule
    \end{tabular}
    \label{tab:zeroshot}
\end{table}

%% file: figures/keywords_tab.tex
\begin{table}[t]
    \centering
    \footnotesize
        \caption{\footnotesize Keywords found by generated features using $\LsWGAN$ and $\LsWGANZ$ for zero-shot ICD-9 codes. Bold words are the most related ones to the ICD-9 code description.}
    \begin{tabularx}{\textwidth}{l|X|X|X}
    \toprule
Code & Description & Keywords from $\LsWGAN$ & Keywords from $\LsWGANZ$  \\ \midrule
V10.62 & Personal history of myeloid leukemia
& AICD, inferoposterior, cardiogenic, \textbf{leukemia}, silent
& \textbf{leukemia}, Zinc, \textbf{myelogenous}, \textbf{CML}, metastases \\ \midrule
E860.3 & Accidental poisoning by isopropyl alcohol  & apneic, pulses, choking, substance,  fractures & \textbf{intoxicated}, \textbf{alcoholic}, AST, EEG, \textbf{alcoholism}  \\ \midrule 
956.3 & Injury to peroneal nerve & vault, \textbf{injury}, pedestrian, orthopedics, \textbf{TSICU} & \textbf{injuries}, \textbf{neurosurgery}, \textbf{injury}, \textbf{TSICU}, coma 
\\ \midrule 
851.05 & Cortex contus-deep coma & \textbf{contusion}, \textbf{injury}, \textbf{trauma}, \textbf{neurosurgery}, \textbf{head} & \textbf{brain}, \textbf{head}, \textbf{contusion},  \textbf{neurosurgery}, \textbf{intracranial} 
\\ \midrule
772.2 & Subarachnoid hemorrhage of fetus or newborn & \textbf{subarachnoid}, \textbf{SAH}, neurosurgical, screening & \textbf{subarachnoid},\textbf{hemorrhages}, \textbf{SAH}, \textbf{newborn}, \textbf{pregnancy}
    \\    
    \bottomrule
    \end{tabularx}
    \label{tab:keywords}
\end{table}

%% file: figures/fewshot_tab.tex
\begin{table}[t]
    \centering
    \footnotesize
    \caption{\footnotesize Few-shot ICD coding results. Scores are averaged over 10 runs on different seeds.}
    \begin{tabular}{l|rrrr|rrrr}
    \toprule 
     & \multicolumn{4}{c|}{Micro} & \multicolumn{4}{c}{Macro} \\
    Method &  Pre & Rec & F1 & AUC & Pre & Rec & F1 & AUC \\
    \midrule
    ZAGRNN~\citep{rios2018few} & \textbf{64.00} & 1.27 & 2.48 & 92.11 & 4.15 & 1.23 & 1.90 & 90.99 \\
    ZAGRNN + $\mathcal{L}_\texttt{LDAM}$~\citep{cao2019learning} & 60.53 & 1.82 & 3.53 & 92.10 & 6.29 & 1.80 & 2.80 & 90.74 \\
    ZAGRNN + Meta~\citep{liu2019large} & 48.88 & 6.75 & 11.84 & 92.15 & 16.65 & 6.77 & 9.62 & 90.92
    \\
    \midrule
    $\LsWGAN$~\citep{xian2018feature} & 29.18 & 18.14 & 22.37 & 92.59 & 20.76 & 18.09 & 19.33 & 90.99 \\
    $\LsWGAN$ +  $\mathcal{L}_\texttt{CLS}$ ~\citep{xian2018feature} & 29.18 & 17.67 & 22.01 & 92.54 & 19.88 & 17.62 & 18.68 & 91.01 \\ 
    $\LsWGAN$ +  $\mathcal{L}_\texttt{CYC}$~\citep{felix2018multi}  & 28.82 & 18.43 & 22.48 & 92.57 & 20.39 & 18.28 & 19.28 & 90.96 \\ \midrule
    $\LsWGANZ$ +  $\mathcal{L}_\texttt{CLS}$ & 27.97 & 17.70 & 21.68 & 92.58 & 20.18 & 17.59 & 18.80 & 91.01
    \\
    $\LsWGANZ$ +  $\LsCYC$ & 28.40 & 18.24 & 22.22 & 92.61 & 20.82 & 18.17 & 19.40 & 90.99
    \\\midrule
    $\LsWGAN$ +  $\mathcal{L}_\texttt{KEY}$ (Ours) &  28.97 & 18.31 & 22.44 & 92.62 & 20.92 & 18.24 & 19.49 & \textbf{91.05} \\ 
    $\LsWGANZ$ (Ours) & 27.66 & 18.81 & 22.39 & 92.56 & 20.45 & 18.81 & 19.59 & 90.97 \\ 
    $\LsWGANZ$ +  $\mathcal{L}_\texttt{KEY}$ (Ours) & 27.95 & \textbf{18.96} & \textbf{22.60} & \textbf{92.63} & \textbf{21.55} & \textbf{18.92} & \textbf{20.15} & 91.00 \\
    \bottomrule
    \end{tabular}
    \label{tab:fewshot}
\end{table}

%% file: conclusion.tex
\section{Conclusion}
We introduced the first feature generation framework, \model, for generalized zero-shot multi-label classification in 
clinical text domain. 
We incorporated the ICD tree hierarchy to design GAN models that significantly improved zero-shot ICD coding without compromising the performance on seen ICD codes. 
We also qualitatively demonstrated that the generated features using our framework can preserve the class semantics as well as the ICD hierarchy compared to existing feature generation methods.
In addition to zero-shot codes, we showed that our method can improve the performance on few-shot codes with limited amount of labeled data.

%% file: appendix.tex
\section{Appendix: Gated Recurrent Units}
Below is the detailed construction of \texttt{GRUCell} in \eqref{eq:grnn} from Section~\ref{sec:base}:
\label{apx:gru}
\begin{align*}
    z^t_l &= \sigmoid(W_z\cdot h^t_l + U_z\cdot \clf_l^{t-1} + b_z) \\
    r^t_l &= \sigmoid(W_r\cdot h^t_l + U_r\cdot \clf_l^{t-1} + b_r) \\
    \clf_l^t &= (1-z^t_l)\odot \clf_l^{t-1} + z^t_l\odot \mathrm{tanh}(W_h\cdot h^t_l + U_h \cdot (r^t_l\odot \clf^{t-1}_l) + b_h)
\end{align*}
where $\odot$ is the dimension-wise multiplication.

\section{Appendix: Generative adversarial networks}
\label{apx:related}


GANs~\citep{goodfellow2014generative} have been extensively studied for generate highly plausible data. The idea of GAN is to train a generator and a discriminator through a minimax game.
The generator takes in a random noise and generate fake data to fool the discriminator while  the discriminator tries to distinguish between generated data and real data.  
The training procedure of GANs can be unstable, thus~\cite{arjovsky2017wasserstein}
proposes Wasserstein-GAN (WGAN) to counter the instability problem by optimizing the Wasserstein distance instead of the original Jenson-Shannon divergence. 
\cite{gulrajani2017improved} further improves WGAN by using gradient instead of weight clipping for the required 1-Lipschitz constraint in WGAN discriminator.

\section{Appendix: More training details}
\label{appendix:base}
\paraname{ICD-9 code information.} We extract the ninth version of the ICD code descriptions and hierarchy from the CDC website\footnote{\url{https://www.cdc.gov/nchs/icd/icd9cm.htm}}. In addition to the official description, we extend the descriptions with medical knowledge, including synonyms and clinical information, crawled from online resources\footnote{\url{http://www.icd9data.com/}}.

\paraname{ZAGRNN.} For the ZAGRNN model, we use 100 convolution
filters with a filter size of 5. We use 200 dimensional word vectors pretrained on PubMed corpus\footnote{\url{https://github.com/ncbi-nlp/BioWordVec}}~\citep{zhang2019biowordvec}.  We use dropout on the word embedding layer with rate 0.5. We
use the ADAM~\citep{kingma2014adam} for optimization
with a minibatch size of 8 and a learning rate
of 0.001.  The final feature size and GRNN hidden layer size are both set to 400. We train the ZAGRNN model for 40 epochs.

\paraname{Meta-embedding.} \cite{liu2019large} proposed meta-embedding for solving large long-tail problem by transferring knowledge from head classes to tail classes. The method naturally fits ICD coding due to the long-tailed code distribution. To apply meta-embedding in ICD coding, we first construct a set of centroids $M$ as the mean of $\feat_l$ for each code $l$ from the training data. Let $\odot$ denote dimension-wise multiplication,  then the meta-embedding for $\feat$ is calculated as:
\begin{align}
\feat^{meta}= \feat + e(\feat)\odot(o(\feat)^\top \cdot M)
\end{align}
where $o(\feat)$ is the attention scores for selecting  centroids $M$ and $e(\feat)$ is a dimension-wise coefficient for selecting the attended features. Both $o$ and $r$ are parameterized as neural networks and are learned during fine-tuning. The final classification is performed by $\predy_l=\sigmoid(\clf_l^\top\cdot\feat_l^{meta})$.

For meta-embedding, we fine-tune the neural network modules $e$ and $o$ using ADAM and set learning rate to 0.0001 and batch size to 32. 

\paraname{WGAN-GP with classification loss.}
\cite{xian2018feature} proposed to add a cross-entropy loss during training WGAN-GP to generate features being correctly classified as conditioned labels. In ICD coding, this loss translates to enforcing $\fakefeat$ being classified as positive for code $l$:
\begin{align}
    \LsCLS = -\log P(y_l=1|\fakefeat) = -\log\sigmoid(\clf_l^\top\cdot\fakefeat_l)
\end{align}

\paraname{WGAN-GP with cycle consistency loss.} Similar to adding $\LsCLS$ to prevent the generated features being random, \cite{felix2018multi} proposed to add a loss that constrains the synthetic representations to generate back their original semantic features. Let $R:\sF\mapsto\sC$ be a linear regression estimate
the label embedding $c_l$ from the generated feature $\fakefeat_l$, the cycle consistency loss is defined as:
\begin{align}
    \LsCYC = ||c_l - R(\fakefeat_l)||^2_2
\end{align}